\documentclass[letterpaper, 10 pt, conference,]{ieeeconf}

\IEEEoverridecommandlockouts                              
\usepackage{graphicx}
\usepackage{siunitx}
\usepackage[utf8]{inputenc}
\usepackage[T1]{fontenc}
\usepackage{amsfonts}
\newcommand{\fig}[1]{Fig.~\ref{#1}}

\usepackage{epsfig} 
\usepackage{amsmath} 
\usepackage{amssymb}  
\usepackage{subcaption}
\usepackage{caption}
\usepackage{color}
\usepackage{verbatim}
\usepackage{wasysym}
\usepackage[table,xcdraw]{xcolor}
\usepackage{graphicx}
\usepackage{gensymb}
\usepackage{xcolor}
\usepackage{latexsym}
\usepackage{multirow}
\usepackage{amsmath}
\usepackage{booktabs,caption}
\usepackage[noadjust]{cite}
\usepackage[flushleft]{threeparttable}
\usepackage[table,xcdraw]{xcolor}

\usepackage[normalem]{ulem} 
\usepackage{wasysym}

\newcommand{\RNum}[1]{\uppercase\expandafter{\romannumeral #1\relax}}
\makeatletter
\newlength\tmp@\newlength\t@mp
\newcommand{\comp}[3]
  {\mathop{ \settowidth\tmp@{$\displaystyle\mathop{#1}^{#3}_{#2}$}
  \hbox to \tmp@{\hss \settowidth\t@mp{$\displaystyle #1$}\setlength\t@mp{.45\t@mp}
  $\displaystyle\mathop{#1}^{\hspace\t@mp #3}_{\hspace{-\t@mp}#2}$
  \hss} }}

\makeatother
\title{\LARGE \bf
Buoyant Choreographies: Harmonies of Light, Sound, and Human Connection\\
}

\author{Dennis Hong$^{1}$ and Yusuke Tanaka$^{1}$
\thanks{$^{\dagger}$ All authors are with the Department of Mechanical and Aerospace Engineering, University of California, Los Angeles, CA, USA 90095 {\tt\small \{yusuketanaka, dennishong\}@g.ucla.edu.}}}

\begin{document}
\twocolumn[{%
\renewcommand\twocolumn[1][]{#1}%
    \maketitle
    \begin{center}
        \centering
        \includegraphics[width=0.99\textwidth, trim={0cm 0 40cm 0},clip]{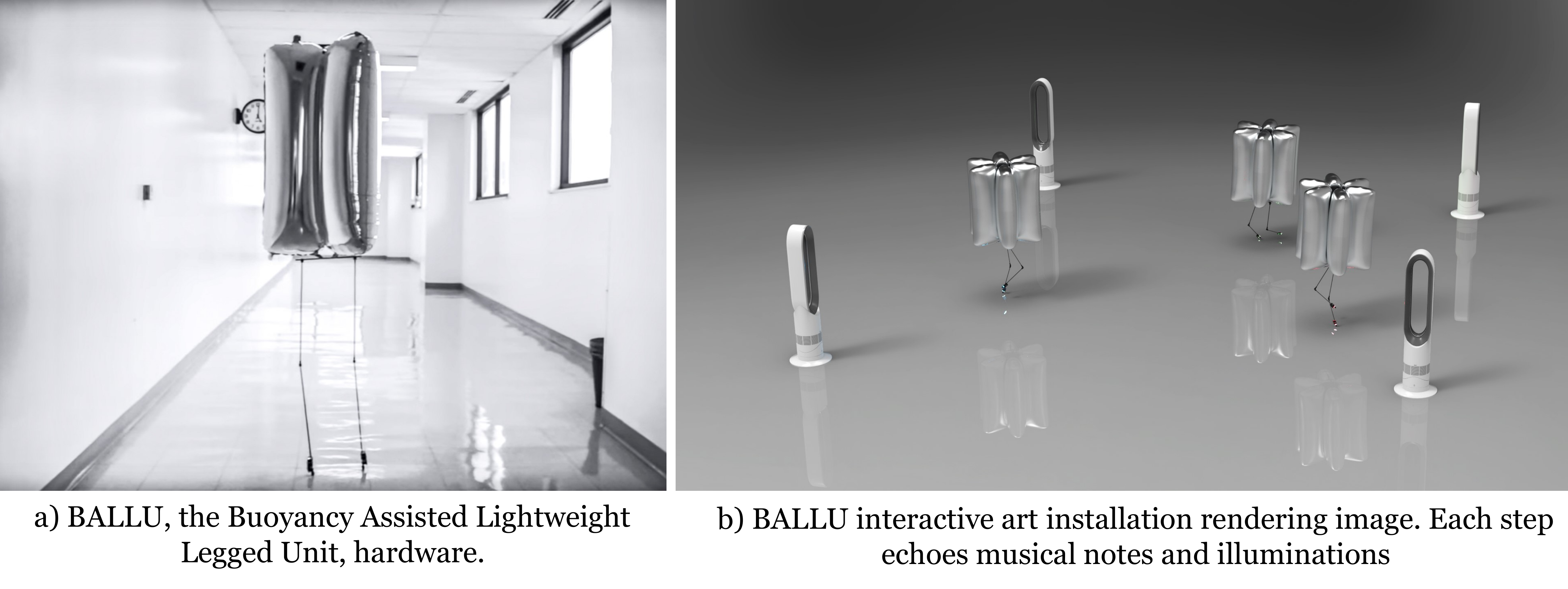} %
    \captionof{figure}{BALLU hardware and interactive art exhibition concept.}
    \label{fig:fig1}
    \end{center}%
    }]

    \footnotetext[1]{All authors are with the Department of Mechanical and Aerospace Engineering, University of California, Los Angeles, CA, USA 90095 {\tt\small \{dennishong, yusuketanaka\}@g.ucla.edu.}}%
\thispagestyle{empty}
\pagestyle{empty}

\section{Abstract}
\subsection{Concept}
BALLU, the Buoyancy Assisted Lightweight Legged Unit \cite{chae2021ballu2}, is a unique legged robot with a helium balloon body and articulated legs \fig{fig:fig1}. Since it is buoyant-assisted, BALLU is inherently stable, never falling over, while being able to walk, jump, and interact safely with people.
The BALLU art installation builds on this playful platform to express fluidity, serendipity, and connection. 
It transforms robotic motion into an artistic visual and acoustic experience, as shown in \fig{fig:hard}, merging technology and creativity into a dynamic, interactive display. This exhibition intentionally does not have a physical boundary for the robots, as shown in \fig{fig:fig1}b, emphasizing the harmony of the technologies and humanity. 
This work significantly extends BALLU's existing permanent exhibition in the Seoul Robotics \& Artificial Intelligence Museum, Seoul RAIM (https://anc.masilwide.com/2261) in \fig{fig:museum}, emphasizing the harmony of robotics and humanity through visual, acoustic, and physical expression.

\begin{figure}[t!]
\centering
\includegraphics[width=0.99\linewidth, trim={0cm 0.5cm 0 1cm},clip]{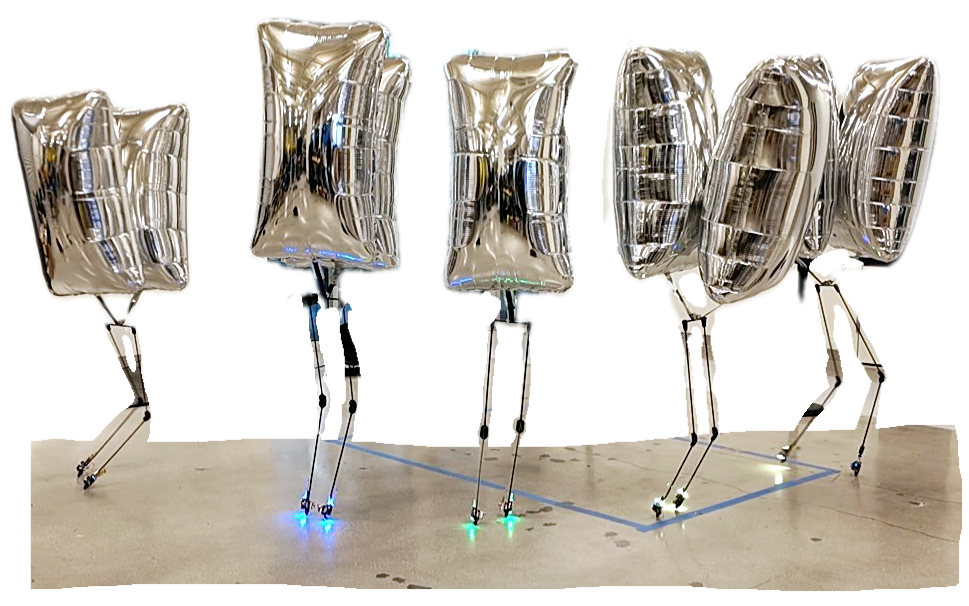}
\caption{Five BALLU robots walking around.}
\label{fig:many}
\end{figure}

\begin{figure}[t!]
\centering
\includegraphics[width=0.7\linewidth, trim={0cm 0.5cm 0 1cm},clip]{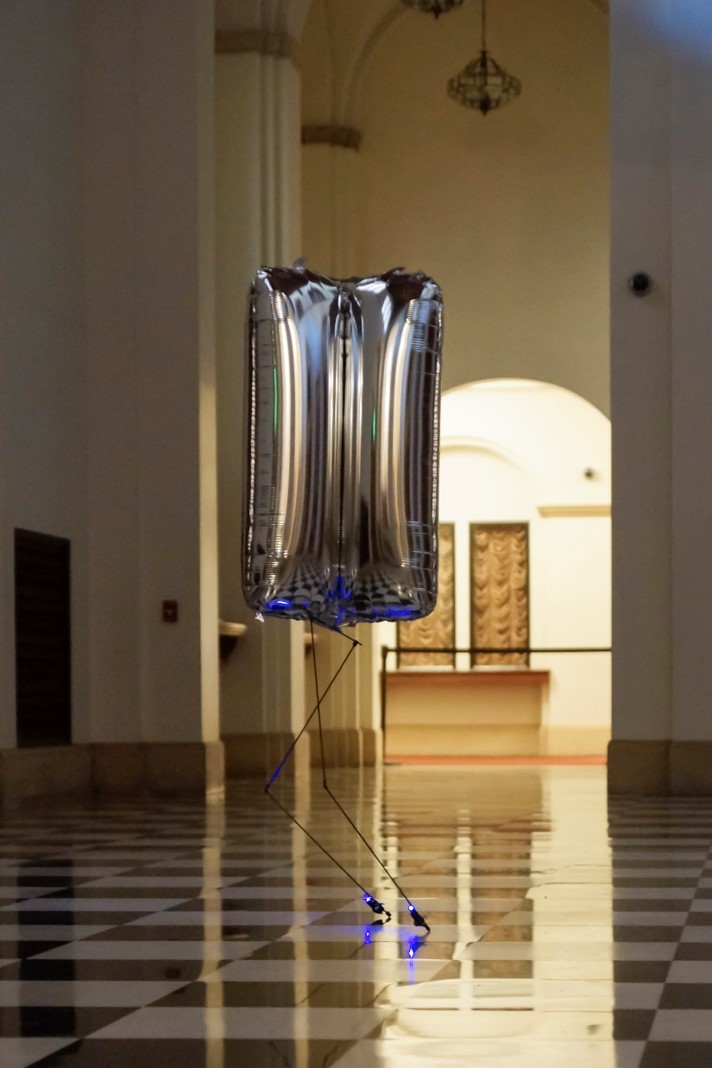}
\caption{BALLU hardware with stepping illumination.}
\label{fig:hard}
\end{figure}

\subsection{Exhibition and Installation Setup}
The installation comprises multiple BALLU robots moving freely within an exhibition area, surrounded by fans that define their motion space and set invisible boundaries as rendered in \fig{fig:spread}. The robots' RGB leg lighting in \fig{fig:hard} changes dynamically with every step, accompanied by musical notes, creating a harmonious blend of motion and melody. Additional fans pointing upwards are placed, introducing a 3D dynamic motion to the exhibit. 

At the end of each motion cycle, the fans converge their airflow toward the center of the field, directing the robots to the center as illustrated in \fig{fig:gather}. This gathering symbolizes unity and interdependence.

\subsection{Human-Art Interaction}
The BALLU art installation invites visitors to interact with the robots directly and indirectly, fostering a deeply engaging experience. People can take control of the robots' movements using gamepads, orchestrating their desired steps and rhythms. Alternatively, physical interactions—such as gently touching the robots or redirecting the airflow of fans—trigger spontaneous and unpredictable reactions to the performance. 

Visitors are encouraged to introduce disturbances to the robots, influencing their gait and melody. These interactions lead to a constantly evolving symphony of stepping notes, creating a dynamic and distinctive auditory and visual experience. Thanks to the stable nature of this balloon robot, it is safe for both visitors and the robots.

The colorful lighting on their legs and synchronized soundscapes transform robots into performers, where the audience can observe and reflect on the interplay between humans, technology, and art.

\begin{figure}[t!]
\centering
\includegraphics[width=0.9\linewidth, trim={12cm 0 0 0},clip]{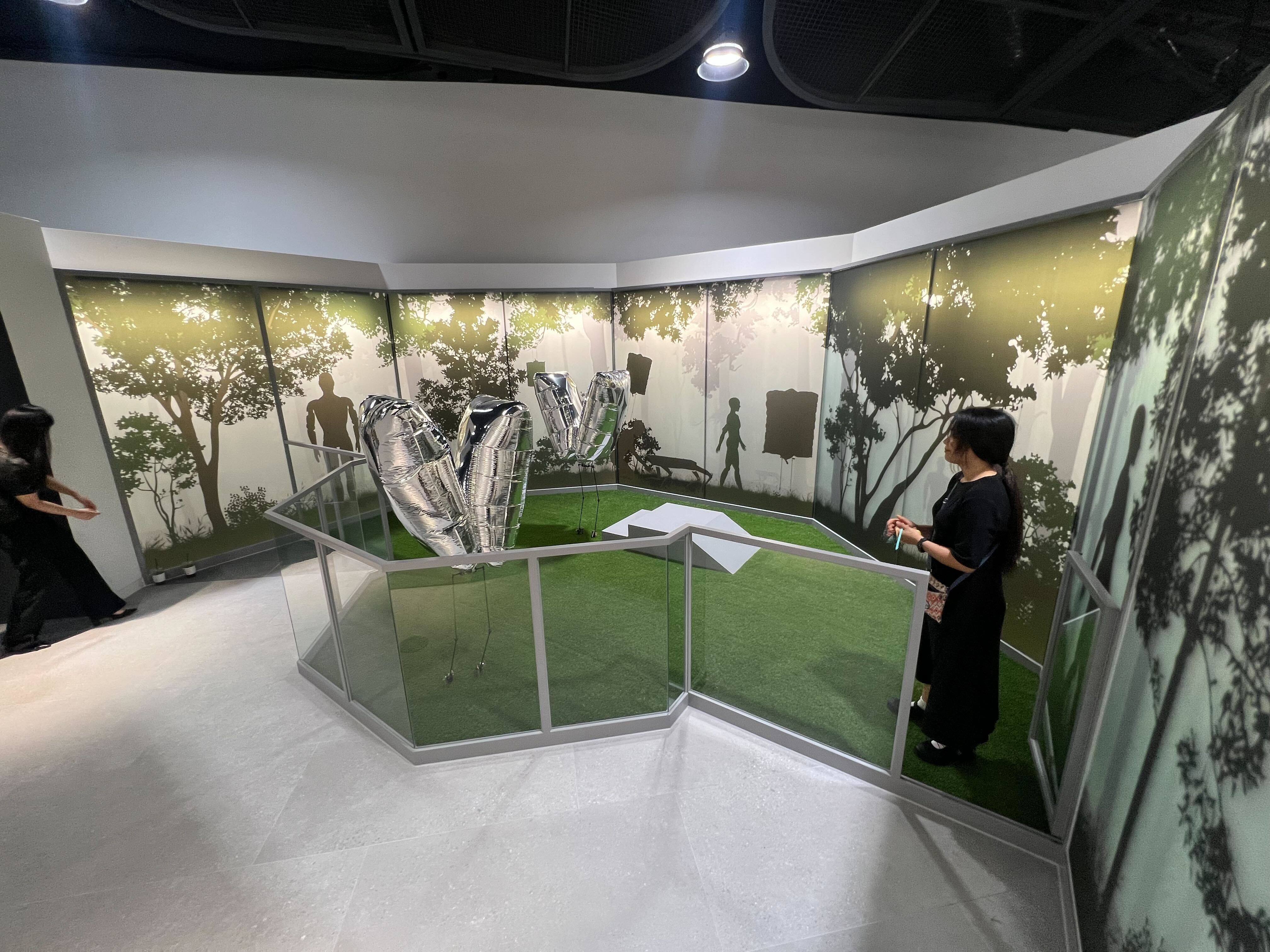}
\caption{BALLU permanent exhibition in Seoul Robotics \& Artificial Intelligence Museum (Seoul RAIM, https://anc.masilwide.com/2261).}
\label{fig:museum}
\end{figure}


\subsection{Technical Details}
The BALLU robots consist of a helium-filled balloon for buoyancy and articulated legs, each of which is controlled by an independent microcontroller with integrated sensors. The legs communicate wirelessly and generate collective motions with the rest of the BALLU legs. This leg-as-a-robot setup poses unique benefits and scientific challenges. Although BALLU is inexpensive and lightweight, each leg consists of IMU, barometers, temperature, force/contact, and power sensors, with sub-GHz and 2.4GHz wireless communications to avoid interference and improve connection reliability. 
Due to the highly nonlinear dynamics of buoyancy-assisted locomotion, precise control is challenging. BALLU is safe and can never fall, making it an ideal platform for real-world, data-driven approaches. 

BALLU's unpredictability is embraced as a core feature of the installation, making each interaction unique.
Ultimately, the BALLU art installation highlights the interplay of robotics and human engagement but also reflects on the fluid boundaries between technology and art. BALLU embodies the harmony of innovation, expression, and shared experience through its unpredictable yet graceful motions.

\begin{figure}[t!]
\centering
\includegraphics[width=0.9\linewidth]{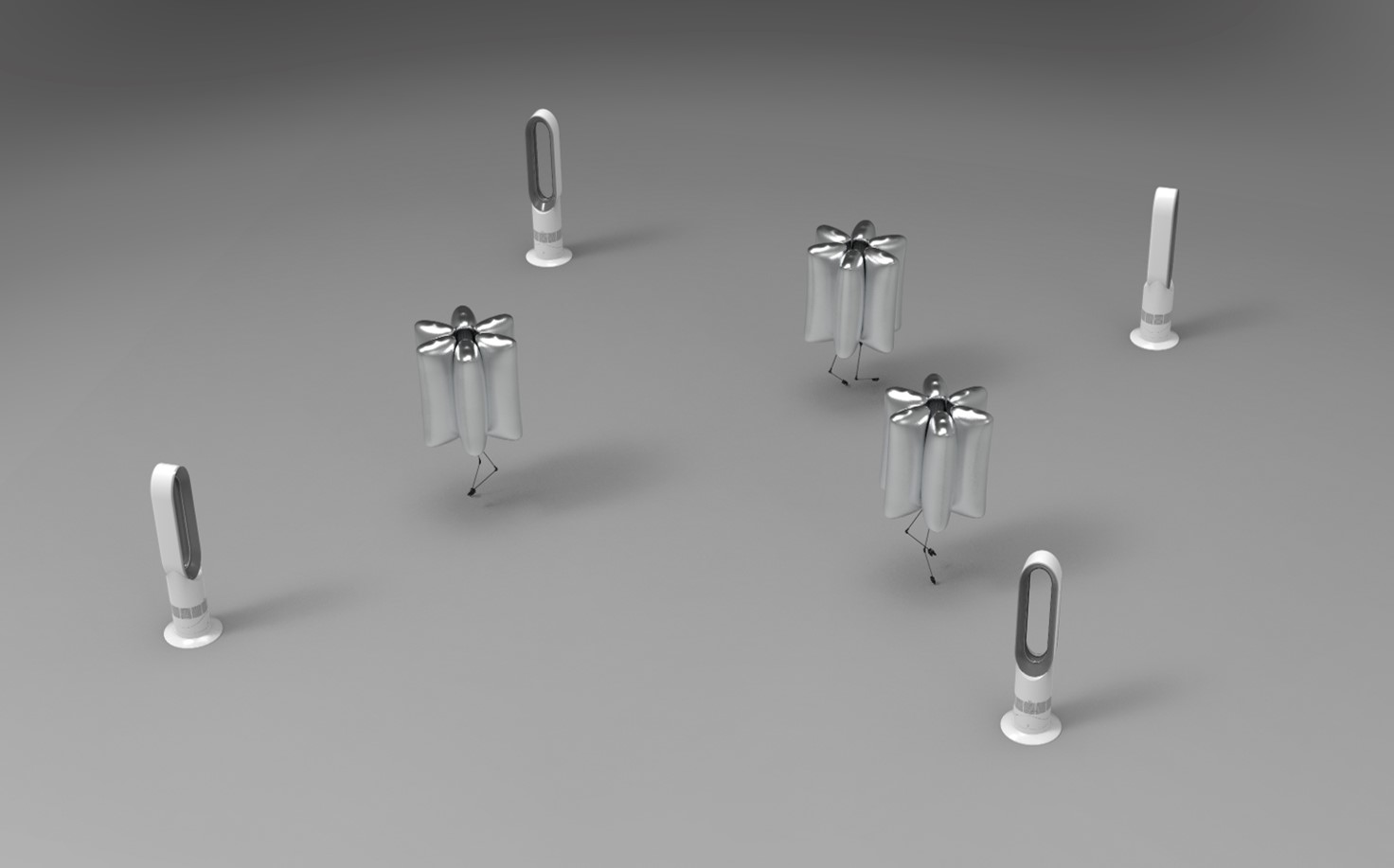}
\caption{Rendering of the BALLU art installation. The four fans at the corners of the arena create an invisible boundary. Visitors are welcome to enter the arena.}
\label{fig:spread}
\end{figure}

\begin{figure}[t!]
\centering
b\includegraphics[width=0.9\linewidth]{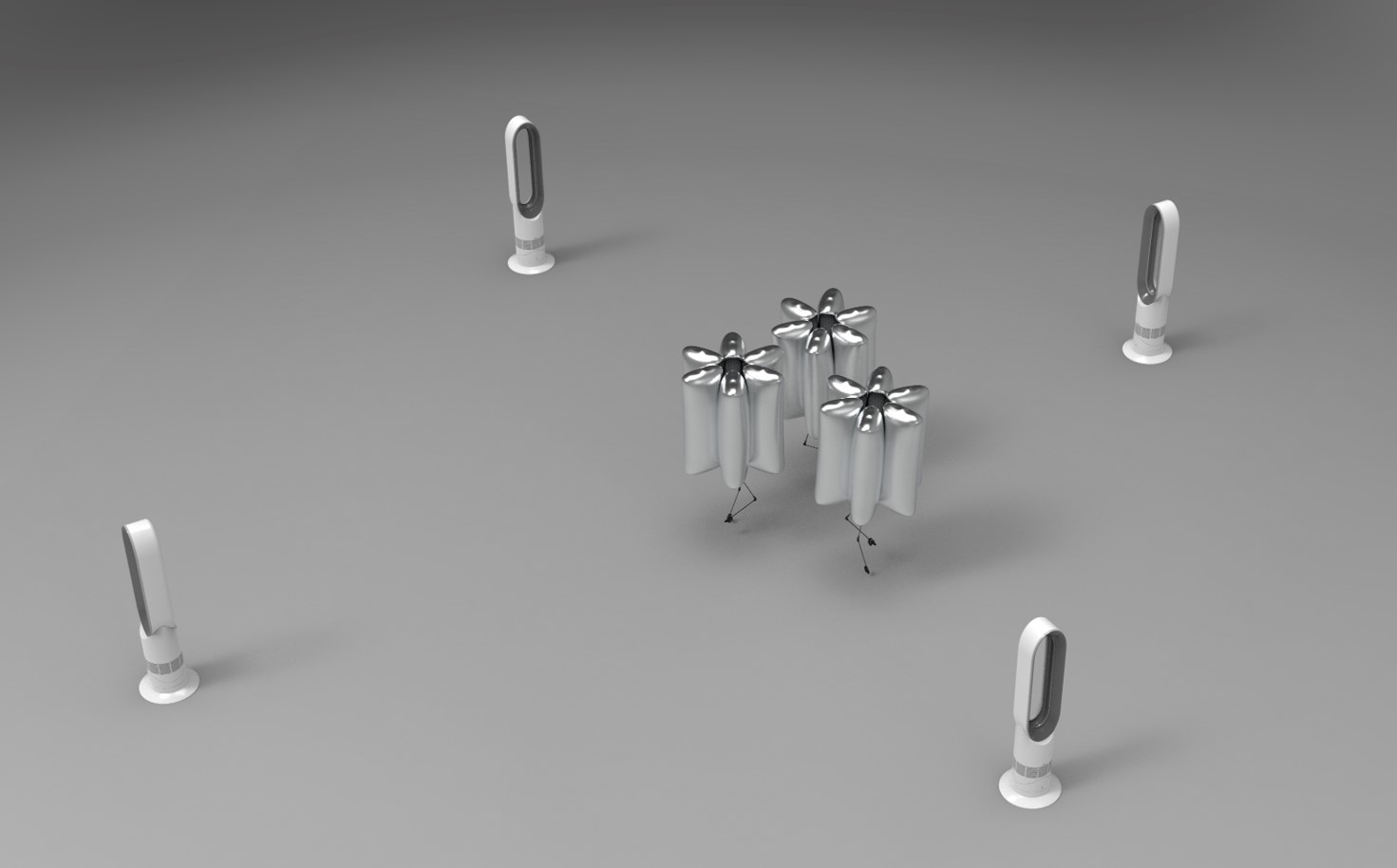}
\caption{Rendering of the BALLU art installation. At the end of the cycle, all the fans point toward the center, bringing BALLU to the center of the arena.}
\label{fig:gather}
\end{figure}

\bibliographystyle{IEEEtran}
\bibliography{main}

\end{document}